# Learnable Image Encryption

Masayuki Tanaka, Member, IEEE

*Abstract--* The network-based machine learning algorithm is very powerful tools. However, it requires huge training dataset. Researchers often meet privacy issues when they collect image dataset especially for surveillance applications. A learnable image encryption scheme is introduced. The key idea of this scheme is to encrypt images, so that human cannot understand images but the network can be train with encrypted images. This scheme allows us to train the network without the privacy issues. In this paper, a simple learnable image encryption algorithm is proposed. Then, the proposed algorithm is validated with cifar dataset.

## I. INTRODUCTION

A machine learning, or a deep learning is very powerful tool in computer vision applications [1]. One of high potential application filed of the deep learning is surveillance. Huge size of training data set is required for the deep learning to obtain high performance. However, especially for surveillance application, it takes large cost to collect enough training data while keeping privacy of people in the data. Nobody wants to be included their photos into the dataset because developers can directly check anyone's behaviors. We can find a similar situation during surveillance operation. Owner of a security camera can check everything.

One of techniques to keep privacy is an image encryption. The image encryption transforms an original image to the encrypted image which people cannot recognize contents in the original image. Image encryption algorithms have been mainly developed to securely transmit images through a public network [2][3]. If people and/or machine want to recognized the contents in the encrypted image, the image is need to be decrypted at first. However, once the image is decrypted, anyone can recognize the contents. It means that the image decryption has possibility to violate the privacy.

In this paper, a novel scheme of the image encryption and the machine learning application is introduced. The key idea is to encrypt images only for human not for machine. In the proposed scheme, the network is directly learned with encrypted images, without decrypting images. Effectiveness of the proposed scheme is demonstrated with well-known cifar dataset.

## II. PROBLEM SETTING

Let's consider two scenarios: training phase and operating phase. In both scenarios, the network usually requires plain image dataset. As shown in Fig. 1 (a), the network is usually trained with the plain images. Even if the image dataset is encrypted as shown in Fig. 1 (b), the original plain images should be decrypted to train the network. Here, the person who trains the network is called trainer. The trainer is often different from the data holder who has the training dataset. Then, the data holder cannot provide the dataset to the trainer with those two existing schemes, because providing the plain image dataset violates a privacy policy of the data holder.

The situation in the operating phase is similar to that in the training phase. The network requires the plain image to detect or to classify an object. Even if the encrypted images are stored in the surveillance system, the images should be decrypted for the network. In this sense, the operator, who operates surveillance system with the network, can always check the original plain images.

For those reasons, here, a novel image encryption problem is introduced. The main difference from the existing image encryption problem is desirable properties of the encrypted images. In the existing image encryption problem, the algorithm should encrypt images against human and network. In the image encryption problem introduced in here, it is desirable that encoded images are encrypted for human, while the network can be trainable with encoded images. We call this kind encryption learnable image encryption. The learnable image encryption can encrypt images for human. It means that the data holder can provide their own dataset without violating the privacy policy. Trainers can directly train with encrypted images. It is very helpful to develop the networks, because the data holder and the trainer can avoid the privacy issues. The learnable image encryption works in the operation phase as well. The network is trained with encrypted images. Therefore, the network can detect or classify object with directly encrypted images without decrypting original plain images.

## III. PROPOSED LEARNABLE IMAGE ENCRYPTION

A block-wise pixel shuffling algorithm is proposed for the learnable image encryption. The procedure of the proposed block-wise pixel shuffling algorithm for 8-bit RGB image can be summarized as follow:

1. The 8-bit RGB image is divided MxM-sized blocks.
2. Each block is split to the upper 4-bit and the lower 4-bit images. Then, we have 6-channel image blocks.
3. Intensities of randomly selected pixel position are reversed.
4. Random pixel shuffle is applied.
5. Encrypted image is restored.

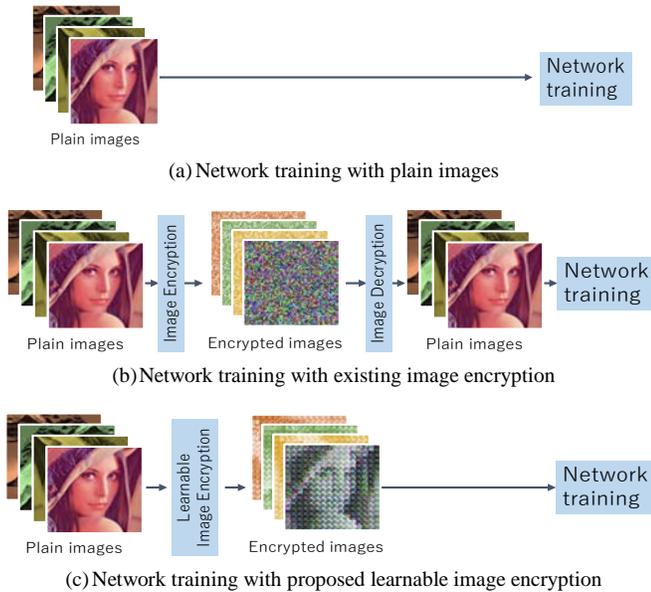

(a) Network training with plain images

(b) Network training with existing image encryption

(c) Network training with proposed learnable image encryption

Fig. 1. Three different types of network training.

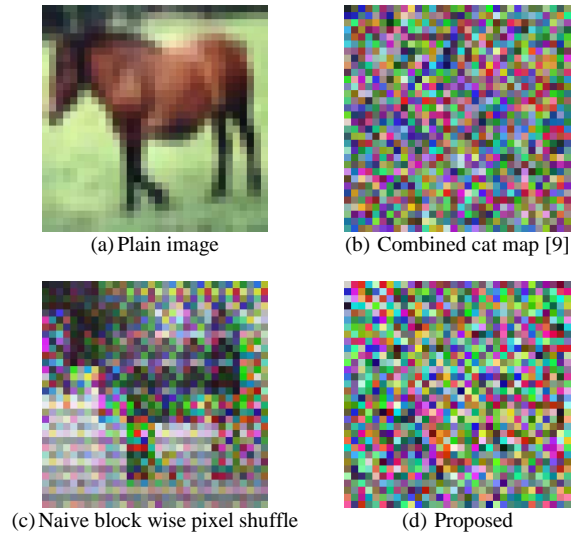

(a) Plain image  (b) Combined cat map [9]

(c) Naive block wise pixel shuffle  (d) Proposed

Fig. 2. Image encryption results.

TABLE I
VALIDATION ACCURACIES OF CIFAR DATASET

|  | CIFAR 10 | CIFAR 100 |
| --- | --- | --- |
| PLAIN IMAGE | 0.884 | 0.591 |
| COMBINED CAT MAP [9] | 0.468 | 0.209 |
| NAIVE BLOCK WISE PIXEL SHUFFLE | 0.872 | 0.602 |
| PROPOSED | 0.863 | 0.568 |

The network structure should adjust to the proposed learnable image encryption. However, the adaptation is simple. A convolution layer with MxM-sized filter and MxM stride is put for the first layer, in order to handle block-wise image encryption. After stacking several network-in-network style layers [4], the feature map is up-sampled to original-sized resolution by the sub-pixel convolution [5]. After this block-wise adaptation, any kind of network can be followed.

## IV. EXPERIMENTS

Experiments were conducted with cifar dataset [6]. The image size of cifar dataset is 32x32. The block-size of the proposed algorithm is set to four. A pyramidal residual networks were built after block adaptation networks [7][8]. Here, plain images, the image encryption based on combined cat map [9], a naive block wise pixel shuffle, and the proposed algorithm are compared[†].

Figure 2 shows an example of image encryption results. The combined cat map and the proposed algorithm can encrypt, while one can recognize a boundary of object from the result of the naive block wise pixel shuffle. Validation accuracies of trained networks with each encryption algorithm are summarized in Table I. The higher validation accuracy represents the better results. The accuracy of the combined cat map algorithm is very worse, while those of other three algorithms are comparable. Although the encrypted image by the proposed algorithm looks random as same as that by the combined cat map, the proposed algorithm can keep the validation accuracies as similar as those of plain images. Those experimental comparisons validate that the only proposed algorithm can encrypt images to satisfy the desired properties of the learnable image encryption.

## V. CONCLUSION

A novel concept of learnable image encryption has been introduced. The learnable image encryption encodes images only for human, but not for the machine. It is expected that the learnable image encryption allows us to develop the network without privacy issues.

REFERENCES

[1] Y. LeCun, Y. Bengio, G. Hinton, Deep learning, nature, vol. 521, no. 7553, pp. 436-444, 2015.
[2] K. Kurihara, M. Kikuchi, S. Imaizumi, S. Shiota, and H. Kiya, An encryption-then-compression system for JPEG standard, IEICE Transactions on Fundamentals of Electronics, Communications and Computer Sciences E98.A(11), pp. 2238-2245, 2015.
[3] S. Assad, M. Farajallah, and C. Valdeanu, Chaos-based block ciphers: An overview, IEEE International Conference on Communications (COMM), pp. 1-4, 2014.
[4] M. Lin, Q. Chen, and S. Yan, Network in network, International Conference on Learning Representations (ICLR), 2014.
[5] W. Shi, J. Caballero, F. Huszar, J. Totz, A. Aitken, R. Bishop, D. Rueckert, and Z. Wang, Real-time single image and video super-resolution using an efficient sub-pixel convolutional neural network, IEEE Conference on Computer Vision and Pattern Recognition (CVPR), pp. 1874--1883, 2016.
[6] A. Krizhevsky, Learning multiple layers of features from tiny images, Tech Report, 2009.
[7] Y. Yamada, M. Iwamura, and K. Kise, Deep pyramidal residual networks with separated stochastic depth, arXiv preprint arXiv:1612.01230, 2016.
[8] D. Han, J. Kim, and J. Kim, Deep pyramidal residual networks, arXiv preprint arXiv:1610.02915, 2016.
[9] X. Wang, L. Liu, and Y. Zhang, A novel chaotic block image encryption algorithm based on dynamic random growth technique, Optics and Lasers in Engineering, vol. 66, pp. 10--18, 2015.

[†] The code is available online.
http://www.ok.sc.e.titech.ac.jp/~mtanaka/proj/imagescramble/